\documentclass{article}
\usepackage{arxiv}
\usepackage{cite}
\usepackage[utf8]{inputenc} 
\usepackage[T1]{fontenc}    
\usepackage{hyperref}       
\usepackage{url}            
\usepackage{booktabs}       
\usepackage{amsfonts}       
\usepackage{nicefrac}       
\usepackage{microtype}      
\usepackage{graphicx}
\usepackage{doi}
\usepackage{amsmath,amssymb,amsfonts}
\usepackage{algorithmic}
\usepackage{textcomp}
\usepackage{xcolor}
\usepackage{subfigure}
\usepackage{hyperref}
\title{High Performance Across Two Atari Paddle Games Using the Same Perceptual Control Architecture Without Training}
\author{Tauseef~Gulrez \\
Department of Computing and Research,\\
Syscon Australia Pty. Ltd.\\
Melbourne, VIC, Australia,\\
\texttt{gtauseef@ieee.org}
\And
Warren~Mansell\\
Division of Psychology and Mental Health,\\
School of Health Sciences, \\ University of Manchester, UK,\\ \texttt{warren.mansell@manchester.ac.uk}
}
\renewcommand{\shorttitle}{\textit{High Performance on Atari using PCTagent}}

\begin{document}

\maketitle

\begin{abstract}
Deep reinforcement learning (DRL) requires large samples and a long training time to operate optimally. Yet humans rarely require long periods training to perform well on novel tasks, such as computer games, once they are provided with an accurate program of instructions. We used perceptual control theory (PCT) to construct a simple closed-loop model which requires no training samples and training time within a video game study using the Arcade Learning Environment (ALE). The model was programmed to parse inputs from the environment into hierarchically organised perceptual signals, and it computed a dynamic error signal by subtracting the incoming signal for each perceptual variable from a reference signal to drive output signals to reduce this error. We tested the same model across two different Atari paddle games Breakout and Pong to achieve performance at least as high as DRL paradigms, and close to good human performance. Our study shows that perceptual control models, based on simple assumptions, can perform well without learning. We conclude by specifying a parsimonious role of learning that may be more similar to psychological functioning.  
\end{abstract}
%
\section{Introduction}\label{sec:introduction}
Gaming environments are increasingly used to develop artificial intelligence. In order to play in a gaming environment, deep reinforcement learning (DRL) agents require a long training time to learn and execute commands. The most successful model-free DRL agents DQN~\cite{mnih2015human}, A3C~\cite{mnih2016asynchronous}, and Rainbow~\cite{hessel2018rainbow} which achieved human-level of performance on Atari games, go through a trial-and-error training process for 10 days (approx. 20 Million steps). Moreover, attempts have been made at developing self-playing agents for Atari games~\cite{chiappa2017recurrent, kaiser2019model}, but none of them were able to achieve high performance. More recently, MuZero agent ~\cite{schrittwieser2020mastering} shows that planning can achieve high performance on Atari games but it requires an extensive engineering cost involving a high computational budget. MuZero agent requires more than 2 months of training to train one agent. In this letter, we propose a new gaming Agent called PCTagent. It is arguably the first agent to achieve human-level performance on the Atari ball and paddle games by exhibiting a control architecture similar to the living organisms.
\begin{figure}[ht]
\centering
\subfigure[]{\includegraphics[width=0.25 \columnwidth]{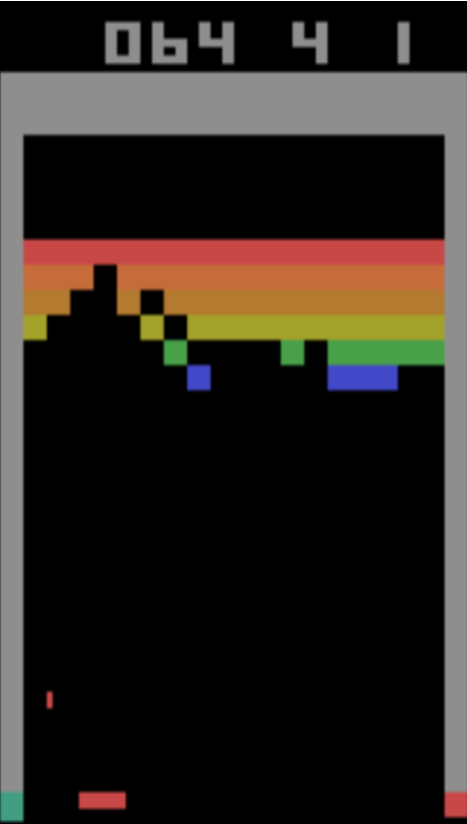}}
\subfigure[]{\includegraphics[width=0.265 \columnwidth]{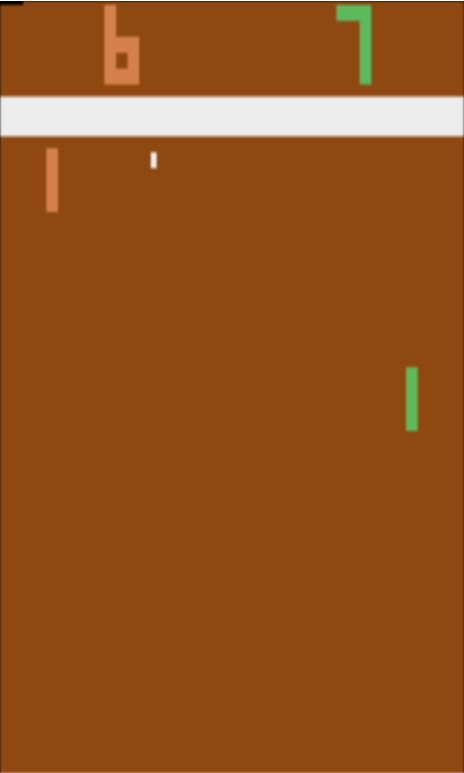}}
\caption{OpenAI Gym's Atari game environments. (a) Breakout and (b) Pong's rgb frames.}
\label{fig:games}
\end{figure}
\subsection{Perceptual Control Theory Modelling}
Perceptual Control Theory (PCT) is a computational framework~\cite{powers1960general,powers1960general2, powers1973behavior, powers2008living} to explain and model the behavior of living organisms based on control engineering. Within a PCT system, output (efferent) signals from each control unit within one level of a hierarchy set the goal state for input (afferent) signals at the level below. This allows actions to vary dynamically to control input 'on the fly', so that the planning and learning of actions is unnecessary in many circumstances. A simple, but 'correct', architecture is necessary to achieve a good level of performance. The nervous system of a living organism requires the input functions to construct perceptual signals from the environment, and to organise these with respect to one another. This architecture may have biologically prepared foundations~\cite{plooij2020phylogeny}, and be further organised through verbal instruction within humans~\cite{brown2018perceptual}; both of these pathways bypass long periods of training.

A variety of PCT-based computational models have been designed that do not require training, which have emulated naturalistic behaviour within animals~\cite{bell2015evolving}, human manual tracking~\cite{parker2020systematic}, human crowd behaviour~\cite{mcphail1992simulating}, flyball catching~\cite{marken2005optical} and robotic devices~\cite{young2017general},~\cite{Barter2021.01.22.427862}. However, there is little published evidence using established benchmarks within commonly used hardware or modelling environments to compare with approaches that do require training for the learning of actions. This was the aim of the current study.
\subsection{Existing AI Methods for Arcade Games}
Atari arcade games have been benchmarked primarily using model-free RL algorithms. DQN~\cite{mnih2015human} utilises deep neural network (DNN) to train Q-learning policies by incorporating replay experience and target networks. Several attempts have been made to extend DQN by incorporating bias correction, e.g. DDQN ~\cite{van2016deep}, and by prioritising experience replay~\cite{schaul2015prioritized} by architectural modifications ~\cite{wang2016dueling}, and distributional value learning~\cite{dabney2018distributional}. 
%
Some attempts have been made to improve performance by data collection, which increases the cost of environment steps beyond 200 million ~\cite{mnih2016asynchronous, badia2020agent57}. Agents developed on proprioceptive inputs~\cite{higuera2018synthesizing,henaff2017model}, model images without using them for planning ~\cite{oh2015action, chiappa2017recurrent, babaeizadeh2017stochastic}, or combine the benefits of model-based and model-free approaches~\cite{kalweit2017uncertainty, nagabandi2018neural}. 
Most model-based agents with image inputs have thus far been limited to relatively simple control tasks~\cite{watter2015embed, hafner2019dream}. 

SimPLe agent~\cite{kaiser2019model} learns a video and predicts a model in pixelated data format and utilises its predictions to train a proximal policy agent~\cite{schulman2017proximal}. The model tracks and establishes prediction based on four consecutive frames and incorporates discrete latent variable as an input. The authors evaluate SimPLe on a subset of Atari games for 400k and 2M environment steps, after which the rewards decreased, hence the model over-fitted the environment. 
As a direct contrast to the above designs, we have used perceptual control theory as a computational framework to show a competitive performance with no training. PCTagent was built on a single GPU Core i7 laptop, for the paddle operated game environment, within which it outperforms top single-GPU Atari agents Rainbow~\cite{hessel2018rainbow} and IQN~\cite{dabney2018distributional} which rest upon years of model-free RL research~\cite{fortunato2017noisy}.
\begin{table}\label{tab:comp}
	\centering
	\caption{Ball and Paddle games results of different ML Agents}
\begin{tabular}{lllr}\hline
\textbf{Agent Name} & \multicolumn{2}{c}{\textbf{Best Score}} & \textbf{Training Frames} \\
\textbf{} & \textbf{Breakout} & \textbf{Pong}	& \textbf{} \\ \hline
DQN      & 401   & 21   & 10 million    \\
DBA3C    & 301   & 21   & 100 million \\
SimPLe   & 16.4   & 5.2   & 4 million  \\   
RADAR    & 600+   & 21    & 200 million  \\
Rainbow         & 120    & 21   & 200 million   \\
IQN             &  79    & 21   & 200 million   \\
DreamerV2       & 312    & 21   & 200 million   \\
Human World record    & 864   & 21 & - \\
Random          &   2   & -20 & - \\
\textbf{PCTagent} (Ours)        & 862   & 18* & 0 \\ \hline
*Frameskip version
\end{tabular}
\end{table}

\section{Method}
\subsection{A PCT Agent and Model}
\begin{figure}[h]
\centering
\includegraphics[width=0.5 \columnwidth]{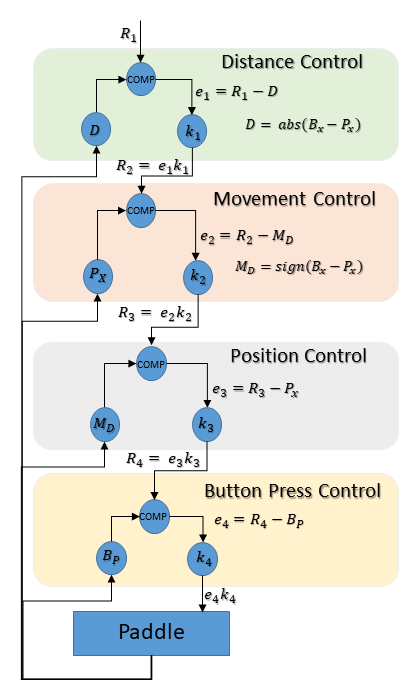}
\caption{Hierarchical PCT model for ball and paddle Atari Games.}
\label{fig:pctagent}
\end{figure}

PCTagent was developed based on four sub-system hierarchical model of behavior, as shown in Fig.~\ref{fig:pctagent} and is available to download under GPL \footnote{\url{https://github.com/PCT-Models/PCTagent_Breakout_Atari}}

The design of a PCT agent is based on a systematic logical analysis of the sensors and effectors of a system, and the performance requirements of the task, that can be specified as an algorithm~\cite{hawker2020robots}. Within the Breakout scenario, the sensor can estimate the distance between the paddle and ball, which needs to be controlled at zero to perform the task. Yet the effector only commands the rate of button press left or right. Therefore, a hierarchy was constructed that bridges between the control of button press at the bottom level and the control of paddle-ball distance at the top level. Two intermediate levels were required. The full architecture can therefore be described as follows.    
The top level of PCTagent controls the visual perception of the distance ($D$) to be maintained between the paddle and a moveable ball in the game environment, as shown in Fig.~\ref{fig:games}(a)(b). Within Fig.~\ref{fig:pctagent}, the error ($e_1$) in the top sub-system sets the reference value ($R_2$) for the direction control left or right of the paddle, which is compared to the sensed direction of the paddle ($M_D$), to generate an error ($e_2$) that sets the reference value ($R_3$) for the next level down. The next level down perceives the the position of the paddle ($P_x$) and compares this to the reference value for movement ($R_3$). Error in this system ($e_3$) sets the reference value ($R_4$) for the left or right button press ($B_P$), and in turn the error of this system ($e_4$) is transformed to a frequency of button press which determines how far left or right user has to move to hit the ball. An inherent embedded property of button press sub-system is the control of the velocity at which the paddle has to move to catch the ball. Within this sub-system, rate of button press is counted by an integrator, which has a limit (currently set to 3), upon reaching max, the integrator resets and reverts to its previous position. Each level of the model contains a $k$ parameter which is the gain of each unit. The gain values for this PCT agent were all set to the value of 1 because this required no a priori assumptions regarding their optimal value.
\subsection{ALE Environment and Image Processing}
We used the Arcade Learning Environment (ALE) as a platform to empirically assess the performance of our PCTagent. ALE provides an interface for different Atari 2600 game environments which are challenging and engaging for humans. ALE serves as a benchmark environment for the evaluation of AI agents. As an input to our PCTagent we obtained raw Atari frames, which are 210X160 (height,width) pixel image with a 128 color representation. To make the game computationally efficient, we reduced the input matrix dimensionality by first converting it to binary and then cropping it to obtain a 100X132 region, capturing only the playing area. We do not require any specific square size images, mostly required by 2D convolution algorithms. The tracking of the ball and the paddle were done using simple contour detection function of OpenCV2 module running under Python 3.8.
%
\section{Atari Paddle Games Experiments}
First we present our PCTagent's performance results and a comparison of this performance with other state-of-the-art agent as tabulated in Table-1. In Table-1, it can be seen that all of the machine learning game playing agents require millions of frames for training for weeks, whereas since PCTagent does not require any training, hence was ready to play instantly. PCTagent's highest score for Breakout and Pong was 862 and 18, respectively.

\begin{figure}[h]
\centering
\subfigure[]{\includegraphics[width=0.6 \columnwidth]{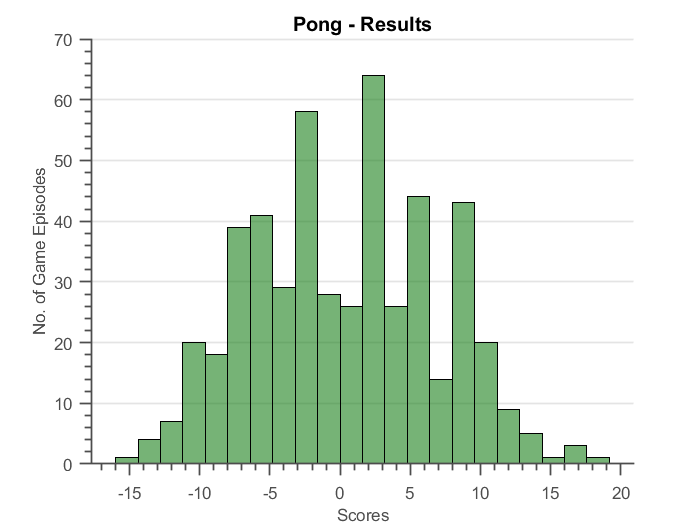}}
\subfigure[]{\includegraphics[width=0.6 \columnwidth]{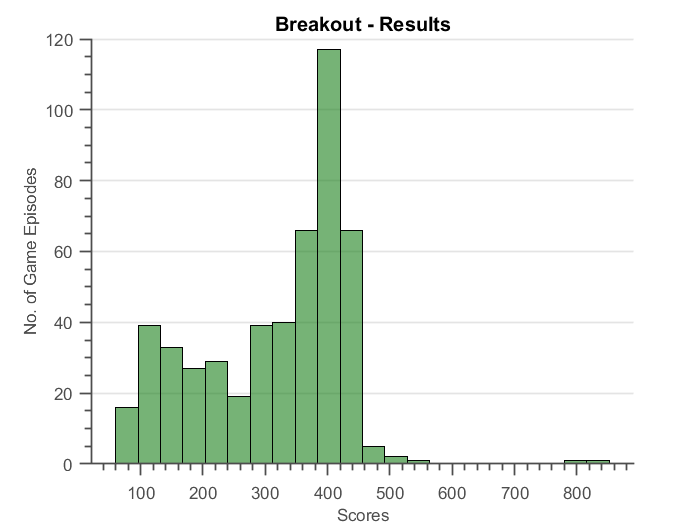}}
\caption{Histogram score representation of the 500 episodes of (a) Pong and (b) Breakout games.}
\label{fig:scores}
\end{figure}

Paddle games e.g. Breakout and Pong were used to evaluate the performance of our training free PCTagent. The games and PCTagent were run on an Intel Core-i7 single GPU machine with AMD Radeon RX 640 graphics card. Both Pong and Breakout were tested for different Open Gym environment modes i.e. with/without frame skipping and deterministic mode. The games were ran for 500 episodes and scores data were collected as shown in Fig.~\ref{fig:scores}. In Breakout, a negative reward of -1 for each life lost and in Pong a -1 reward for opponent's successful score was set. It is evident from the results (shown in Fig.~\ref{fig:scores}) that in Breakout it is possible to achieve an average highest score of 400+ and in Pong to win the game with a margin of 5-6 points, at no prior training costs. In Pong, the noframeskip mode resulted in all wins - scores of 21, hence frameskip mode was used during data collection. A significant noise was introduced in the PCTagent when the game were stuck in one position. In the case of Breakout this usually happened in the end (noframeskip mode) when one or two bricks were left to hit, which resulted most of the time in losing a life.
\section{Conclusion}
We produced one controller based on PCT and tested it, without training, on two different Atari paddle games. Its performance matched or exceeded that of published benchmarks from existing reinforcement and deep learning models. These findings complement earlier studies demonstrating the high performance of PCT controllers within robotics and other areas of research~\cite{young2017general},~\cite{Barter2021.01.22.427862}. Indeed, the findings are particularly consistent with an earlier comparison between a PCT controller and an LQR controller for an inverted pendulum robot~\cite{johnson2020implementation}. This PCT controller required minimal tuning and its performance metrics were superior. 

PCT controllers can provide a robust control solution across environments with no training because they have no need to learn their actions. Instead, they achieve control by varying their outputs on-the-fly to control their inputs by acting against unpredictable disturbances (e.g. turbulence, rough ground), unlike reinforcement models. The closed-loop PCT design emulates control systems in nature, which also do not need to learn specific behaviours to operate effectively and efficiently~\cite{yin2020crisis}. The choice of input specifications and their hierarchical organisation can be made through a systematic analysis of the sensors, effectors and the task requirements rather than through learning, or inferences made by the researcher~\cite{hawker2020robots}. This begs the question of whether reinforcement learning accounts for human skills or whether PCT provides a more accurate model - an architecture of 'priors' proposed to forge future advances in AI ~\cite{bengio2021deep}. Importantly, PCT controllers can further improve performance through training if required. The learning algorithm specified in PCT - reorganisation - uses random-walk learning to optimise the parameters (e.g. gains) and functions (e.g. specification of inputs that co-vary with target velocity) within a PCT architecture. This is a future direction for computational modelling using PCT.
%
%
\bibliographystyle{plain}
\bibliography{references}

\begin{thebibliography}{10}

\bibitem{babaeizadeh2017stochastic}
Mohammad Babaeizadeh, Chelsea Finn, Dumitru Erhan, Roy~H Campbell, and Sergey
  Levine.
\newblock Stochastic variational video prediction.
\newblock {\em arXiv preprint arXiv:1710.11252}, 2017.

\bibitem{badia2020agent57}
Adri{\`a}~Puigdom{\`e}nech Badia, Bilal Piot, Steven Kapturowski, Pablo
  Sprechmann, Alex Vitvitskyi, Zhaohan~Daniel Guo, and Charles Blundell.
\newblock Agent57: Outperforming the atari human benchmark.
\newblock In {\em International Conference on Machine Learning}, pages
  507--517. PMLR, 2020.

\bibitem{Barter2021.01.22.427862}
Joseph~W. Barter and Henry~H. Yin.
\newblock Achieving natural behavior in a robot using neurally inspired
  hierarchical control.
\newblock {\em bioRxiv}, 2021.

\bibitem{bell2015evolving}
Heather~C Bell, Greg~D Bell, Jeffrey~A Schank, and Sergio~M Pellis.
\newblock Evolving the tactics of play fighting: Insights from simulating the
  “keep away game” in rats.
\newblock {\em Adaptive Behavior}, 23(6):371--380, 2015.

\bibitem{bengio2021deep}
Yoshua Bengio, Yann Lecun, and Geoffrey Hinton.
\newblock Deep learning for ai.
\newblock {\em Communications of the ACM}, 64(7):58--65, 2021.

\bibitem{brown2018perceptual}
Carla Brown-Ojeda and Warren Mansell.
\newblock Do perceptual instructions lead to enhanced performance relative to
  behavioral instructions?
\newblock {\em Journal of motor behavior}, 50(3):312--320, 2018.

\bibitem{chiappa2017recurrent}
Silvia Chiappa, S{\'e}bastien Racaniere, Daan Wierstra, and Shakir Mohamed.
\newblock Recurrent environment simulators.
\newblock {\em arXiv preprint arXiv:1704.02254}, 2017.

\bibitem{dabney2018distributional}
Will Dabney, Mark Rowland, Marc Bellemare, and R{\'e}mi Munos.
\newblock Distributional reinforcement learning with quantile regression.
\newblock In {\em Proceedings of the AAAI Conference on Artificial
  Intelligence}, volume~32, 2018.

\bibitem{fortunato2017noisy}
Meire Fortunato, Mohammad~Gheshlaghi Azar, Bilal Piot, Jacob Menick, Ian
  Osband, Alex Graves, Vlad Mnih, Remi Munos, Demis Hassabis, Olivier Pietquin,
  et~al.
\newblock Noisy networks for exploration.
\newblock {\em arXiv preprint arXiv:1706.10295}, 2017.

\bibitem{hafner2019dream}
Danijar Hafner, Timothy Lillicrap, Jimmy Ba, and Mohammad Norouzi.
\newblock Dream to control: Learning behaviors by latent imagination.
\newblock {\em arXiv preprint arXiv:1912.01603}, 2019.

\bibitem{hawker2020robots}
Benjamin Hawker and Roger~K Moore.
\newblock Robots producing their own hierarchies with dosa; the
  dependency-oriented structure architect.
\newblock {\em UK-Robotics and Autonomous Systems (RAS) Network}, pages 66--68,
  2020.

\bibitem{henaff2017model}
Mikael Henaff, William~F Whitney, and Yann LeCun.
\newblock Model-based planning with discrete and continuous actions.
\newblock {\em arXiv preprint arXiv:1705.07177}, 2017.

\bibitem{hessel2018rainbow}
Matteo Hessel, Joseph Modayil, Hado Van~Hasselt, Tom Schaul, Georg Ostrovski,
  Will Dabney, Dan Horgan, Bilal Piot, Mohammad Azar, and David Silver.
\newblock Rainbow: Combining improvements in deep reinforcement learning.
\newblock In {\em Proceedings of the AAAI Conference on Artificial
  Intelligence}, volume~32, 2018.

\bibitem{higuera2018synthesizing}
Juan Camilo~Gamboa Higuera, David Meger, and Gregory Dudek.
\newblock Synthesizing neural network controllers with probabilistic
  model-based reinforcement learning.
\newblock In {\em 2018 IEEE/RSJ International Conference on Intelligent Robots
  and Systems (IROS)}, pages 2538--2544. IEEE, 2018.

\bibitem{johnson2020implementation}
Thomas Johnson, Zhou Siteng, Wei Cheah, Warren Mansell, Rupert Young, and Simon
  Watson.
\newblock Implementation of a perceptual controller for an inverted pendulum
  robot.
\newblock {\em Journal of Intelligent \& Robotic Systems}, 99(3-4):683--692,
  2020.

\bibitem{kaiser2019model}
Lukasz Kaiser, Mohammad Babaeizadeh, Piotr Milos, Blazej Osinski, Roy~H
  Campbell, Konrad Czechowski, Dumitru Erhan, Chelsea Finn, Piotr Kozakowski,
  Sergey Levine, et~al.
\newblock Model-based reinforcement learning for atari.
\newblock {\em arXiv preprint arXiv:1903.00374}, 2019.

\bibitem{kalweit2017uncertainty}
Gabriel Kalweit and Joschka Boedecker.
\newblock Uncertainty-driven imagination for continuous deep reinforcement
  learning.
\newblock In {\em Conference on Robot Learning}, pages 195--206. PMLR, 2017.

\bibitem{marken2005optical}
Richard~S Marken.
\newblock Optical trajectories and the informational basis of fly ball
  catching.
\newblock 2005.

\bibitem{mcphail1992simulating}
Clark McPhail, William~T Powers, and Charles~W Tucker.
\newblock Simulating individual and collective action in temporary gatherings.
\newblock {\em Social Science Computer Review}, 10(1):1--28, 1992.

\bibitem{mnih2016asynchronous}
Volodymyr Mnih, Adria~Puigdomenech Badia, Mehdi Mirza, Alex Graves, Timothy
  Lillicrap, Tim Harley, David Silver, and Koray Kavukcuoglu.
\newblock Asynchronous methods for deep reinforcement learning.
\newblock In {\em International conference on machine learning}, pages
  1928--1937. PMLR, 2016.

\bibitem{mnih2015human}
Volodymyr Mnih, Koray Kavukcuoglu, David Silver, Andrei~A Rusu, Joel Veness,
  Marc~G Bellemare, Alex Graves, Martin Riedmiller, Andreas~K Fidjeland, Georg
  Ostrovski, et~al.
\newblock Human-level control through deep reinforcement learning.
\newblock {\em nature}, 518(7540):529--533, 2015.

\bibitem{nagabandi2018neural}
Anusha Nagabandi, Gregory Kahn, Ronald~S Fearing, and Sergey Levine.
\newblock Neural network dynamics for model-based deep reinforcement learning
  with model-free fine-tuning.
\newblock In {\em 2018 IEEE International Conference on Robotics and Automation
  (ICRA)}, pages 7559--7566. IEEE, 2018.

\bibitem{oh2015action}
Junhyuk Oh, Xiaoxiao Guo, Honglak Lee, Richard Lewis, and Satinder Singh.
\newblock Action-conditional video prediction using deep networks in atari
  games.
\newblock {\em arXiv preprint arXiv:1507.08750}, 2015.

\bibitem{parker2020systematic}
Maximilian~G Parker, Andrew~BS Willett, Sarah~F Tyson, Andrew~P Weightman, and
  Warren Mansell.
\newblock A systematic evaluation of the evidence for perceptual control theory
  in tracking studies.
\newblock {\em Neuroscience \& Biobehavioral Reviews}, 112:616--633, 2020.

\bibitem{plooij2020phylogeny}
Frans~X Plooij.
\newblock The phylogeny, ontogeny, causation and function of regression periods
  explained by reorganizations of the hierarchy of perceptual control systems.
\newblock In {\em The Interdisciplinary Handbook of Perceptual Control Theory},
  pages 199--225. Elsevier, 2020.

\bibitem{powers1973behavior}
W.~T. Powers.
\newblock {\em Behavior: The control of perception}.
\newblock Aldine Chicago, 1973.

\bibitem{powers1960general2}
William~T Powers, RK~Clark, and RL~McFarland.
\newblock A general feedback theory of human behavior: Part ii.
\newblock {\em Perceptual and Motor Skills}, 11(3):309--323, 1960.

\bibitem{powers2008living}
William~Treval Powers.
\newblock Living control systems iii: The fact of control.
\newblock 2008.

\bibitem{powers1960general}
William~Treval Powers, Robert~K Clark, and RL~Mc Farland.
\newblock A general feedback theory of human behavior: Part i.
\newblock {\em Perceptual and motor skills}, 11(1):71--88, 1960.

\bibitem{schaul2015prioritized}
Tom Schaul, John Quan, Ioannis Antonoglou, and David Silver.
\newblock Prioritized experience replay.
\newblock {\em arXiv preprint arXiv:1511.05952}, 2015.

\bibitem{schrittwieser2020mastering}
Julian Schrittwieser, Ioannis Antonoglou, Thomas Hubert, Karen Simonyan,
  Laurent Sifre, Simon Schmitt, Arthur Guez, Edward Lockhart, Demis Hassabis,
  Thore Graepel, et~al.
\newblock Mastering atari, go, chess and shogi by planning with a learned
  model.
\newblock {\em Nature}, 588(7839):604--609, 2020.

\bibitem{schulman2017proximal}
John Schulman, Filip Wolski, Prafulla Dhariwal, Alec Radford, and Oleg Klimov.
\newblock Proximal policy optimization algorithms.
\newblock {\em arXiv preprint arXiv:1707.06347}, 2017.

\bibitem{van2016deep}
Hado Van~Hasselt, Arthur Guez, and David Silver.
\newblock Deep reinforcement learning with double q-learning.
\newblock In {\em Proceedings of the AAAI Conference on Artificial
  Intelligence}, volume~30, 2016.

\bibitem{wang2016dueling}
Ziyu Wang, Tom Schaul, Matteo Hessel, Hado Hasselt, Marc Lanctot, and Nando
  Freitas.
\newblock Dueling network architectures for deep reinforcement learning.
\newblock In {\em International conference on machine learning}, pages
  1995--2003. PMLR, 2016.

\bibitem{watter2015embed}
Manuel Watter, Jost~Tobias Springenberg, Joschka Boedecker, and Martin
  Riedmiller.
\newblock Embed to control: A locally linear latent dynamics model for control
  from raw images.
\newblock {\em arXiv preprint arXiv:1506.07365}, 2015.

\bibitem{yin2020crisis}
Henry Yin.
\newblock The crisis in neuroscience.
\newblock In {\em The Interdisciplinary Handbook of Perceptual Control Theory},
  pages 23--48. Elsevier, 2020.

\bibitem{young2017general}
Rupert Young.
\newblock A general architecture for robotics systems: A perception-based
  approach to artificial life.
\newblock {\em Artificial life}, 23(2):236--286, 2017.

\end{thebibliography}
\end{document}